\documentclass[runningheads]{llncs}
\usepackage[T1]{fontenc}
\usepackage{marvosym}
\usepackage{times}
\usepackage{epsfig}
\usepackage{graphicx}
\usepackage{amsmath}
\usepackage{algorithm}
\usepackage{algorithmicx}
\usepackage{algpseudocode}
\usepackage{color}
\usepackage{enumitem}
\usepackage{arydshln}
\usepackage{multirow}
\usepackage{bm}
\usepackage{latexsym}
\usepackage{booktabs}
\usepackage{amssymb}
\usepackage{hhline}
\usepackage{subcaption}
\usepackage{graphicx}
\begin{document}
\title{Read Pointer Meters based on a Human-like Alignment and Recognition Algorithm}
%
%
\author{Yan Shu\inst{1,2} \and
Shaohui Liu\inst{2} \textsuperscript{\Letter} \and
Honglei Xu\inst{2} \ and
Feng Jiang\inst{2}}
\authorrunning{Yan Shu et al.}
%
\institute{State Key Laboratory of Communication Content Cognition, People's Daily Online, Beijing, 100733, China \and
Computer Science and Technology Department, Harbin Institute of Technology, Harbin, 150001, China
\\
\email{shliu@hit.edu.cn}
}
\maketitle              
\begin{abstract}
Recently, developing an automatic reading system for analog measuring instruments has gained increased attention, as it enables the collection of numerous types of equipment. Nonetheless, two major obstacles still obstruct its deployment to real-world applications. The first issue is that they rarely take the entire pipeline’s speed into account. The second is that they are incapable of dealing with some low-quality images (i.e., meter breakage, blur, and uneven scale). In this paper, we propose a human-like alignment and recognition algorithm to overcome these problems. More specifically, a spatial transformed module (STM) is proposed to obtain the front view of images in a self-autonomous way based on an improved spatial transformer network (STN). Meanwhile, a value acquisition module (VAM) is proposed to infer accurate meter values by an end-to-end trained framework. In contrast to previous research, our model aligns and recognizes meters totally implemented by learnable processing, which mimics human behaviors and thus achieves higher performance. Extensive results verify the robustness of the proposed model in terms of accuracy and efficiency. The code and the dataset is available in https://github.com/shuyansy/Detect-and-read-meters.

\end{abstract}
\begin{figure}
\centering
\begin{subfigure}{0.4\columnwidth}
    \includegraphics[width=\textwidth]{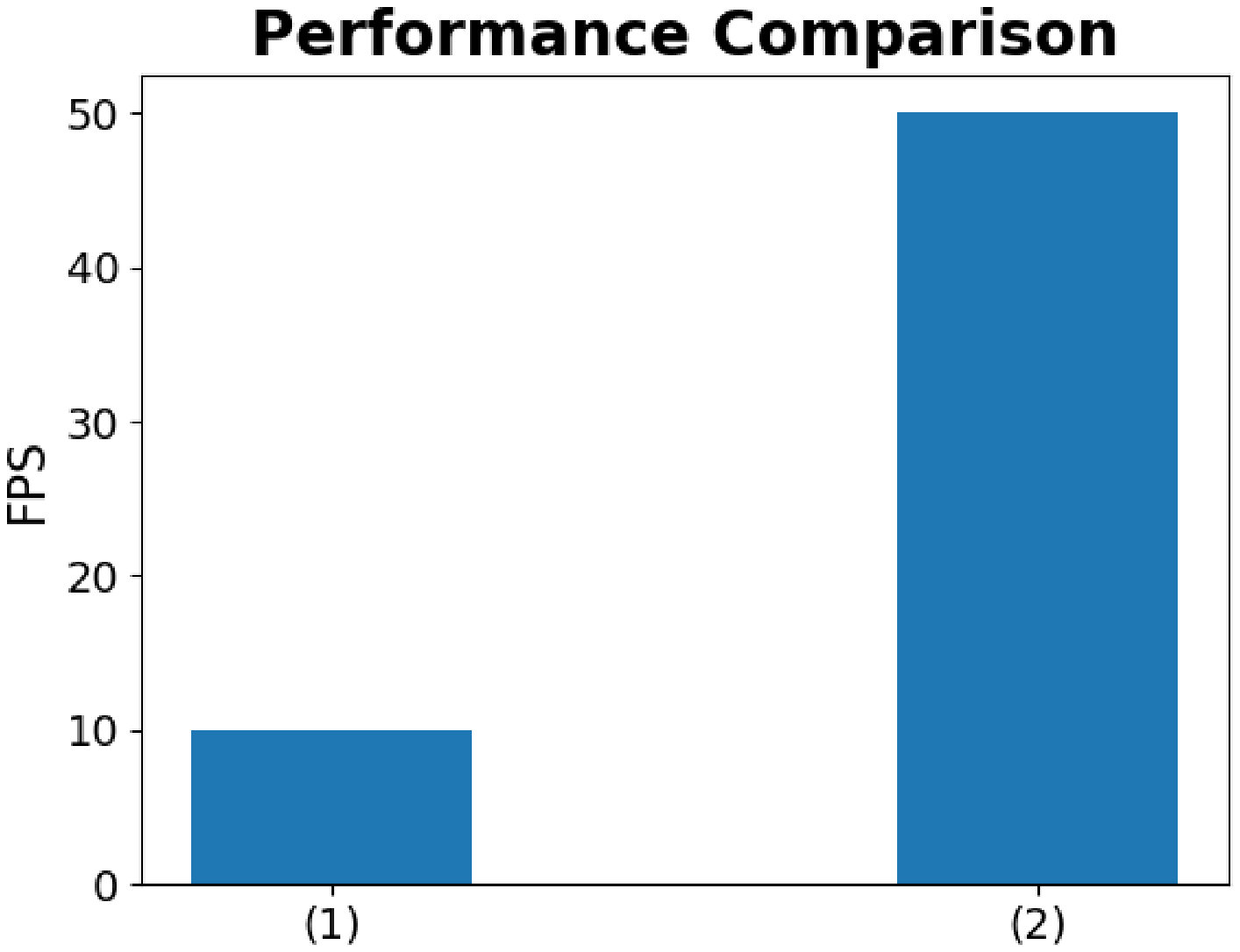}
    \caption{}
    \label{fig:first}
\end{subfigure}
\begin{subfigure}{0.4\columnwidth}
    \includegraphics[width=\textwidth]{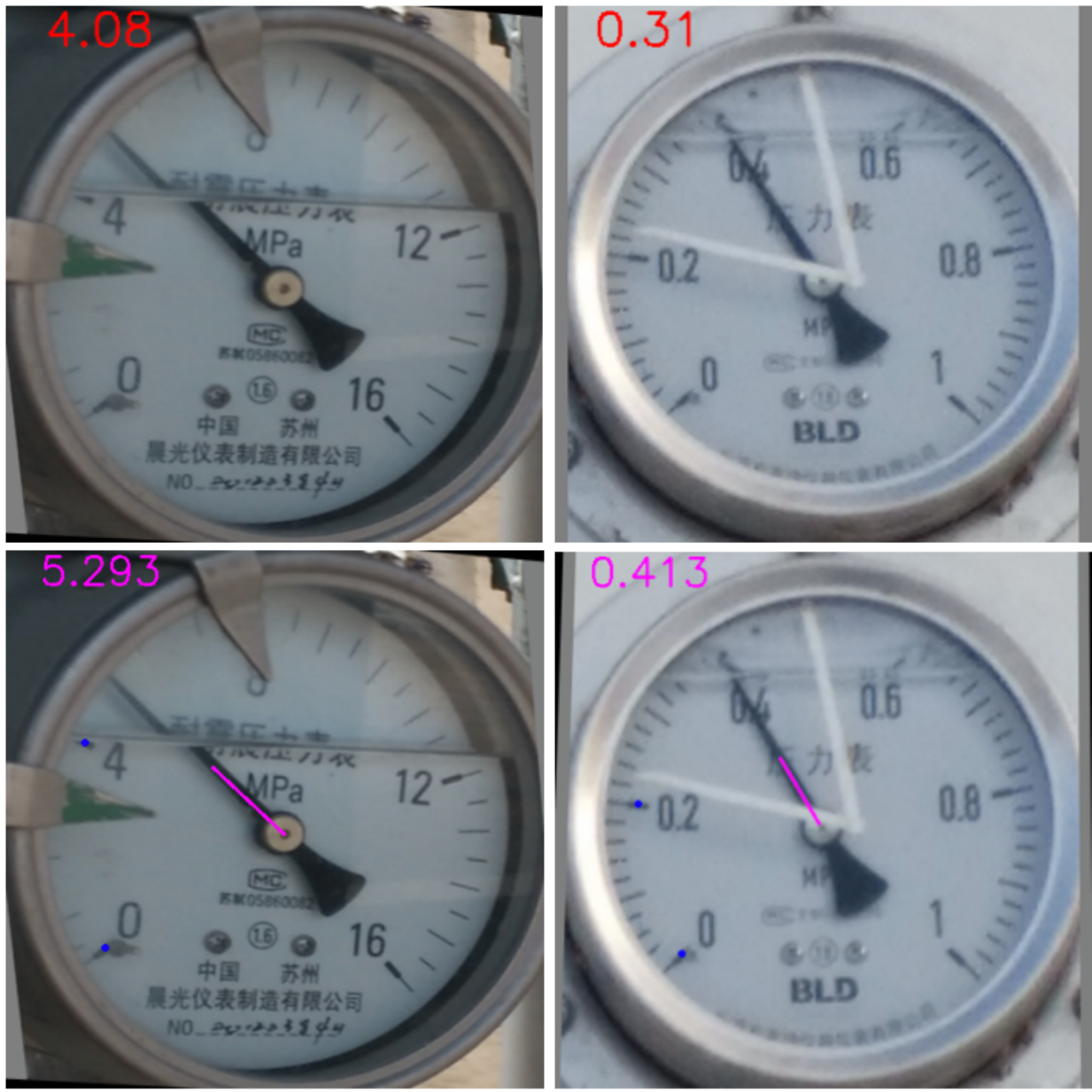}
    \caption{}
    \label{fig:second}
\end{subfigure}

\caption{(a) shows the efficiency of our STM (2) for meter alignment, which is 5 times faster than the conventional perspective transform method (1). (b) shows that our VAM (bottom line) can read more accurate values in some low-quality images than prior methods (top line).}
\label{compare}
\end{figure}

\section{Introduction}
\label{sec:intro}
In the complex industrial environment, there are harsh environments such as radiation, toxicity, and high temperature, and it is necessary to inspect the production conditions with the help of instruments to ensure safety \cite{article1} . Traditionally acquired data are typically read artificially by humans, who are capable of deriving precise readings from complex meters in a variety of shapes, forms, and styles, despite never having seen the meter in question. However, the manual method is always more labor intensive and time consuming. Therefore, it is of great practical significance to rely on inspection robots and computer vision technology \cite{article2,article3,article4,article5} for automatic meter reading.

\begin{figure*}[t]
\begin{center}
\includegraphics[width=0.9\linewidth]{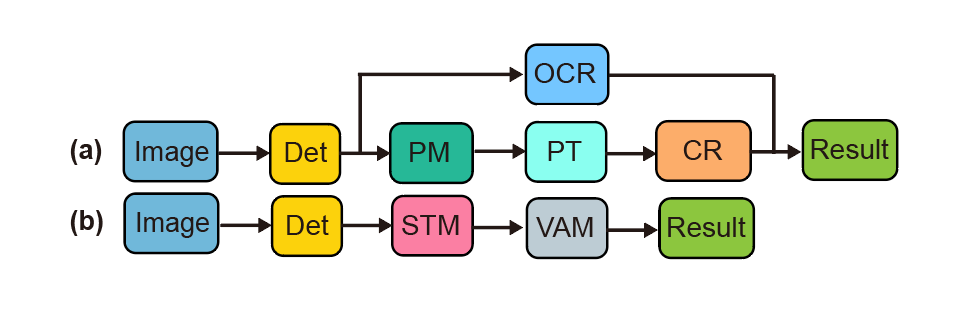}
\end{center}
\vspace{0pt}
\caption{
Overview of previous pointer meter reading pipeline (a) compared to ours(b). ``Det", ``PM", ``PT" and ``CR" represent meter detection, point matching, perspective transform and component retrieval. ``STM" and ``VAM" are our spatial transformed module and value acquisition module.}
\label{figpipeline}
\end{figure*}
Substation meters are now classified as digital and pointers.
While reading digital meters can be considered an OCR task and is relatively simple to accomplish using text spotting techniques \cite{QinICCV,MaskTextspotter,FOTS,ABCNet}, as demonstrated in Appendix A, reading pointer meters presents a different and more difficult problem: there are major visual changes between meter faces, the camera viewpoint has a significant effect on their depicted shape and numbering location, and the existence of shadows, meter breakage, and specular reflections adds to the pointer hands' perplexity. While this issue has been around for a long time, few previous solutions have been capable of reliably obtaining readings from meters, except in extremely limited circumstances. Additionally, it is difficult for researchers to work on this project due to the lack of reliable training and evaluation standards.

Existing automatic meter reading systems \cite{meter1,meter2,meter3,meter4}, according to relevant literature, include the following pipelines. To begin, the meter's pure area is detected using conventional neural network-based detection algorithms or image processing techniques; then, the captured target is aligned to a front view by the perspective transform method. Lastly, meter values can be obtained by meter component (the pointer and the scale) retrieval and meter number recognition. However, most of these methods suffer from two main problems. First, the alignment process is typically time-consuming due to its intricate point-matching steps, which hinders the overall efficiency of the system. Second, their reading model is not robust; it consists of isolated and independent modules for meter component retrieval and number recognition, which are unaware of their interdependence, resulting in poor accuracy. Therefore, ``how to design an algorithm for efficient alignment and robust recognition of pointer meters" remains largely unsolved.

To address these issues, we propose a novel human-like alignment and recognition algorithm that simplifies the meter reading pipeline, as shown in Fig. \ref{figpipeline}. To be more precise, we propose a novel spatial transformed module (STM) for alignment via implicitly learning homography transformation, which is heavily inspired by spatial transformer networks (STN) \cite{stn} . STM is more efficient at aligning meters than previous morphological conversion methods by discarding the point-matching process. Additionally, a value acquisition module (VAM) is established in a unified framework of meter component retrieval and meter number recognition, simulating the structure of an end-to-end text spotter. By excavating the relationship between the meter component and meter number, VAM can learn a richer representation and thus can read precise meter values from low-quality images. As shown in Fig. \ref{compare}, on the MC1260 dataset we proposed, the FPS of STM is 50 FPS, which is 5 times faster than the conventional alignment method. Meanwhile, VAM can handle some difficult data, such as meter breakage, blur and uneven scale.

 In this paper, we make the following contributions:
(i) We design a unified framework involving detection, alignment and recognition stages. The detection can simply be an off-the-shelf object detection model. The alignment stage involves a deep neural network that introduces an improved STN to regress homography transformation parameters implicitly. At the recognition stage, we are the first to establish an end-to-end architecture to tightly couple meter component retrieval and meter number recognition, boosting both the accuracy and efficiency of pointer meter reading.

(ii) We propose a new benchmark dataset called the Meter Challenge (MC1296) which contains 1296 images captured in scene by automatic robots. MC1296 is organized in a tree structure, containing images, annotations and evaluation metrics for different tasks (meter detection, meter alignment, and meter recognition) from top to bottom.

(iii) Extensive experiments verify the effectiveness and robustness of the method we propose.

The rest of this paper is organized as follows. The related background knowledge is provided in Section \ref{sec:re}, including the previous pointer meter reading pipelines, spatial transformer networks (STN) and end-to-end text spotting methods highly related to our work. Section \ref{sec:method} introduces the implementation process of the proposed method. In Section \ref{sec:experiment}, the proposed method is verified by extensive simulation experiments and ablation studies. The conclusions of this paper are summarized in Section  \ref{sec:conclusion}.

\section{Related Works}
\label{sec:re}
We commence this section by reviewing major pointer meter reading frameworks. Additionally, we discuss the research on STN and end-to-end text spotting methods,  which is highly relevant to our work.

\subsection{Pointer meter reading frameworks} 
Numerous advances \cite{meter1,meter2,meter3,meter4,meterdeep,metermaskrcnn,meterrobust,metertemplate}  have been made in the reading of pointer meters over the last few years. The existing frameworks are generally divided into three stages: meter detection, meter alignment, and meter recognition. Traditional algorithms \cite{metertemplate} such as template matching and the table lookup method are used in meter detection. To address this issue with complex backgrounds, some object detection methods, such as Faster R-CNN \cite{meter4} , have been introduced. To calibrate the camera angle to obtain a front view image, perspective transform techniques \cite{meter4,meter2}  are applied by calculating the transformation matrix determined by point matching. Image processing methods \cite{meterdesign}  also propose using the image subtraction method or the Hough transform algorithm to extract the pointer for meter recognition. Additionally, machine learning and deep learning are used to improve reading accuracy. Liu et al. \cite{metersvm}  used SVM to separate meters, while He et al. \cite{metermaskrcnn}  improved the Mask R-CNN \cite{maskrcnn} method for pointer segmentation. Then, the final values can be determined by calculating the pointer angle and meter number output.

The majority of the aforementioned approaches are able to read pointer meters, but few of them can balance accuracy and speed due to complex post-processing in meter alignment \cite{meter4} or inadequate visual representations in meter recognition \cite{metersvm}.

\subsection{Spatial Transformer Networks (STN)} 
  This is in contrast to the conventional perspective transform method, which explicitly calculates the transformation matrix. STN \cite{stn} introduces a novel learnable module that enables spatial manipulation of data within the network. STN is advantageous for a wide variety of computer vision tasks due to its efficiency and flexibility. ASTER \cite{aster} consists of a rectification network and a recognition network that can deal with text that is distorted or has an irregular layout. Lee et al. \cite{imageregis}  propose image-and-spatial transformer networks (ISTNs) for downstream image registration optimization. Additionally, Yang et al. \cite{clock} introduce a clock alignment architecture based on STN, which motivates us to develop a more efficient meter alignment module.

\subsection{End-to-end text spotters}
To spot texts in images, a straight two-stage idea is proposed to cascade existing detectors and recognizers sequentially. However, due to the lack of complementarity between the detector and recognizer, they suffer from low efficiency and accuracy. To mitigate this problem, an end-to-end trainable neural network for text spotting is attempted, with state-of-the-art performances achieved. Li et al. \cite{2017TCRNN}  first built a unified end-to-end work that simultaneously localizes and recognizes text with a single forward pass, with positive results achieved in a horizontal text spotting task. Benefiting from the convolution sharing strategy, FOTS \cite{2018FOTS} and EAA 
 \cite{2018EAA} pool multioriented text regions from the feature map by designing RoI rotate and text-alignment layers, respectively. Unfortunately, few researchers have incorporated end-to-end text spotters into their pointer meter recognition frameworks.

Our work is structured similarly to existing frameworks for pointer meter reading. To increase the applicability of previous work, we replace the traditional perspective transform method with an improved STN and then create an end-to-end meter recognition module for meter component retrieval and meter number recognition.

\begin{figure*}[t]
\begin{center}
\includegraphics[width=\linewidth]{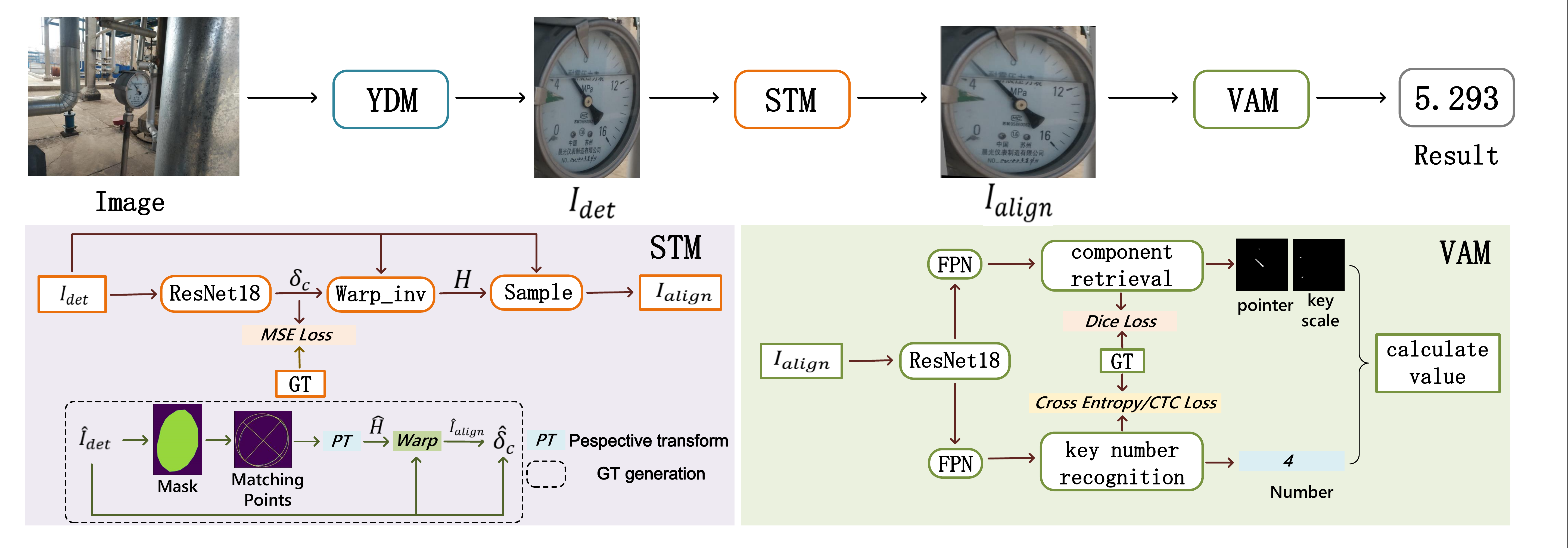}
\end{center}
\vspace{0pt}
\caption{
The proposed framework of the pointer meter recognition. YDM can detect meter targets and crop meter regions into STM, where aligned views can be obtained. VAM can output meter values accurately and efficiently.  
}
\label{figmodel}
\end{figure*}
\section{Methods}
\label{sec:method}
The purpose of this paper is to design an algorithm for the efficient alignment and robust recognition of pointer meters. To achieve this goal, we establish a unified framework, which is shown in Fig. \ref{figmodel}. Our proposed architecture accepts an image as input and then performs detection, alignment, and recognition sequentially. It is worth noting that our STM (see Sec. \ref{MA}) can directly transform the detected meter into an aligned view without any postprocessing steps. Meanwhile, the VAM  (see Sec. \ref{MR}) we proposed can learn rich visual representation by excavating the relationship between component retrieval and number recognition.

\subsection{Meter Detection}
Cropping meter regions prior to recognition is necessary to eliminate background interference. To accomplish this, some traditional image processing techniques, such as Hough circle detection and template matching, are used, both of which have shortcomings in some low-quality images. At the moment, object detection networks are used to detect and crop the meter, as follows.

 \begin{equation}\label{eq1}
\begin{aligned}\
I_{det}=\Phi{_{det}}(I;\Theta {_{det}})\in \mathbb{R}^{^{3 \times h \times w}}
\end{aligned}
\end{equation}

where $I$ is the given unlabelled image, while $\Phi{_{dec}}$ and $\Theta{_{dec}}$ represent the detecting function and learnable parameters, respectively.

The detector, defacto, can be performed using any off-the-shelf object detector. However, to reduce the efficiency cost and handle some small meter targets, we propose a YOLO-based detection module  (YDM) based on YOLO-v5 \cite{yolov4} , which has achieved state-of-the-art performance in many tasks. To achieve better performance in our tasks where data are scarce and the target is small, we apply a multiscale training strategy and artificially augment the images by copy-pasting some small objects. The performance of YDM can be seen in Sec. \ref{sec:experiment}.

\subsection{Meter Alignment}
\label{MA}
\textbf{Motivation.} The detected pure meter image could be directly passed to a module for reading recognition. This is typically not ideal for two reasons: first, due to the limitations of the localization module; and second, even when the meter is properly localized, it can be hard to read at times due to viewpoint interference. Previous methods apply a direct perspective transform to calibrate the camera angle to obtain a front view image, as shown in the following:

 \begin{equation}\label{eq2}
\begin{aligned}\
(x,y,w')=(u,v,w) \cdot T=(u,v,w)\cdot\begin{bmatrix}
a_{11} & a_{12} &a_{13} \\ 
a_{21} & a_{22} &a_{23} \\ 
a_{31} & a_{32} & a_{33}\\
\end{bmatrix}\\
\end{aligned}
\end{equation}

 \begin{equation}\label{eq3}
\begin{aligned}\
(X,Y)=(\frac{x}{w'},\frac{y}{w'})\\
(U,V)=(\frac{u}{w'},\frac{v}{w'})
\end{aligned}
\end{equation}

where $(U,V)$ represents the coordinate of a point in the original image, $(X,Y)$ is the coordinate of the corresponding point in  the transformed image, and $(u,v,w)$ and $(x,y,w')$ are the homogenous space representations of $(U,V)$ and $(X,Y)$, respectively. By matching four feature points between two images, the transform matrix $T$ is determined. Their methods, however, suffer primarily from complex point matching algorithms, which are time-consuming and not very robust.This drives us to design a more efficient and stronger module for meter alignment.

\textbf{Revisiting vanilla STN.} Different from the perspective transform which calculates the transformation matrix by point matching, the STN can transform the detected meter to a fronto-paralleled view by learned homography transformation parameters. Specifically, given the output $I_{det}$ of YDM, STN establishes mapping by predicting homography transformation $H$ with 8 degrees of freedom, and $\phi_{sam}$ represents the Differentiable Image Sampling (DIS) operation to obtain the canonical view of  $I_{det}$ by bilinear interpolation:          

 \begin{equation}\label{eq4}
\begin{aligned}\
H=\Phi_{stn}(I_{det})\in \mathbb{R}^{3\times3} \\
I_{align}=\Phi_{sam}(I_{det},H)\in \mathbb{R}^{3\times h\times w} 
\end{aligned} 
\end{equation}

Therefore, how to predict accurate homography transformation $H$ is a key issue.  

\textbf{Spatial Transformed Module (STM).} It is a direct idea to regress $H$ given ground truth $\hat{H}$ in a supervised way. Nonetheless, based on our major findings and rigorous testing, the deep network fails to learn the explicit parameter of H for the following reasons: (i) The training data are limited to the deep CNN's huge parameters; (ii) $H$'s parameters have a large range of values, making the regression difficult to optimize.
To circumvent these problems, 
we model the implicit spatial transformation relationship between images instead of regressing $H$ directly.    

Specifically, for a $\hat{I}_{det}$ in the training set, we first annotate its inner dial region with a binary mask map. Then, for various meter forms, we match four pairs of feature points to determine the real $\hat{H}$. For an irregular ellipse, the endpoints of the major axis and minor axis are utilized as the initial points, while the corresponding points are defined by the intersection of the major axis, the minor axis, and the circumcircle. For a rectangular shape, the $\hat{H}$ can be calculated by mapping the vertices of the rectangle directly to the vertices of the image. Then, we can obtain the aligned image $\hat{I}_{align}$ by perspective transform:
 
\begin{equation}\label{eq5}
\begin{aligned}\\
\hat{I}_{align}=warp(\hat{I}_{det},\hat{H}) 
\end{aligned}
\end{equation}

The vertex coordinate offsets $\hat{\delta}_{c}$ between $\hat{I}_{det}$ and $\hat{I}_{align}$ can be obtained, which is the training objective of STM implemented by mean-squared (MSE) loss:

 \begin{equation}\label{eq6}
\begin{aligned}\\
{L_{align}}=\sum_{i}(\delta_{ci}-\hat{\delta}_{ci})^2 
\end{aligned}
\end{equation}

where $i$ is the index of coordinates. Therefore, the STM algorithm can be adjusted as follows:

\begin{equation}\label{eq7}
\begin{aligned}\
\delta_{c}=\Phi_{stm}(I_{det})\in \mathbb{R}^{4\times2}\\
H=warp\_inv(I_{det},I_{det}+\delta_{c})\in \mathbb{R}^{3\times3}\\
I_{align}=\Phi_{sam}(I_{det},H)\in \mathbb{R}^{3\times h \times w} 
\end{aligned}
\end{equation}

In our training process, we use ResNet18  \cite{resnet} to extract the feature of $I_{det}$, and by the propagation of the network, accurate $H$ and canonical images can be acquired.

\subsection{Meter Recognition}
\label{MR}
\textbf{Overall Design.} What is the best way to read meters like a human? Key meter elements such as the pointer, scales, and number were predicted in previous methods to achieve this goal. However, they tended to create independent modules to handle different components and numbers, resulting in a suboptimal solution for meter recognition. We propose a unified framework called the value acquisition module (VAM) that consists of a meter component retrieval branch and meter number recognition branch to excavate a deep relationship between them. As illustrated in Fig. \ref{figmodel}, we apply ResNet18 as the backbone and create two separate feature merging modules to form a pair of complementary branches. Specifically, upsampling and pixelwise addition are used to fuse intermediate layers of ResNet. VAM allows these two diametrically different tasks to benefit from each other by disentangling weight sharing and introducing a mirror symmetry of FPN \cite{fpn}. Ablation studies are demonstrated in Sec. \ref{sec:experiment}.

\textbf{Meter component retrieval branch.}
We retrieve the meter component (meter pointer and key scales) using semantic segmentation methods that are heavily inspired by the Mask R-CNN \cite{maskrcnn}. The branch generates two 1-channel segmentation maps, namely, the Pointer Map and the Key Scale Map, by performing two distinct $1 \times 1$ convolutional operations on the backbone features. The Pointer Map indicates the location of the meter's pointer, whereas the Key Scale Map indicates its angle. The Pointer Map and Key Scale Map are both trained by minimizing the Dice loss:

\begin{equation}\label{eq3}
\begin{aligned}
\vspace{8pt}
\vspace{8pt}
\vspace{8pt}
L_{pm}&=1-\dfrac{2 \sum_{i} P_{pm}(i) G_{pm}(i)}{\sum_{i} P_{pm}(i)^{2}+\sum_{i} G_{pm}(i)^{2}}\\ 
L_{k s m}&=1-\dfrac{2 \sum_{i} P_{k s m}(i) G_{k s m}(i)}{\sum_{i} P_{k s m}(i)^{2}+\sum_{i} G_{k s m}(i)^{2}}
\end{aligned}
\end{equation}

where $pm$ and $ksm$ represent Pointer Map and Key Scale Map, respectively, and $P_{(\cdot)}(i)$ refers to the value of $i$\textsuperscript{th} pixel in the predicted result, while $G_{(\cdot)}(i)$ refers to the value of pixel $i$\textsuperscript{th} in the GT region.

The final loss for the meter component retrieval branch is a weighted combination of the two maps, balanced by $\lambda \in (0,1)$ as

 \begin{equation}\label{eq8}
\begin{aligned}\
L{_{com}}=\lambda L{_{PointerMap}} + (1-\lambda) L{_{KeyScaleMap}} 
\end{aligned}
\end{equation}

In our experiments, we set $\lambda$ to 0.4, assigning more importance to the key scale map, which is relatively difficult to learn in the training process due to its small spatial occupation.

\textbf{Meter number recognition branch.}
Previous methods recognize numbers in meters with another system, which poses severe memory waste and low efficiency. In our VAM, the meter number recognition branch resembles like the standard text spotters, which is mentioned in Sec.\ref{sec:re}. To further boost the inference speed, we only detect the key number in the meter, the one closest to the number `0', and then recognize it with the assistance of feature sampling.

The key number detection task is deemed a text classification task, in which one convolution is applied to output dense per-pixel predictions of the key number localization. The key number bounding box can be obtained by the minimum bounding rectangle operation. Meanwhile, to overcome the class imbalance problem, we introduce online hard example mining (OHEM) \cite{ohem} to better distinguish between number areas and backgrounds, in which the balanced factor is set to 3 in our work. The set of positive elements selected  by OHEM in the score map is $\omega$, and the loss function for key number detection can be formulated as:

 \begin{equation}\label{eq9}
\begin{aligned}
L{_{num\_det}}=\frac{1}{\parallel\Omega\parallel }\sum_{x\in\Omega}Cross\_Entropy(p_x,p_x^*)\\
=\frac{1}{\parallel\Omega\parallel }\sum_{x\in\Omega}(-p_x^*logp_x-(1-p_x^*)log(1-p_X))
\end{aligned}
\end{equation}

where $\parallel\cdot\parallel$ means the number of elements in a set, and the $p_x$ and $p_x^*$ are the predicted pixel and the ground truth label, respectively.  

The feature sampling layer aims to convert detected feature regions into fixed-size outputs from which an RNN-based sequence recognizer can be established. We introduce RoI rotate in \cite{FOTS} to our work, which can transform the rotated area into a fixed-size region via max-pooling and bilinear interpolation. Similar to but distinguished from STN, RoI rotate gets affine transformation via an unsupervised way, resulting in a more general operation for extracting features for regions of interest. To improve recognition performance, we use only ground truth key number regions during training rather than predicted number regions.

Given the transformed number feature, we first permute key number features $F\in \mathbb{R}^{C\times H\times W}$ into 2D sequence feature $L\in \mathbb{R}^{C\times W}$ in several sequential convolutions, which has the same configurations as CRNN \cite{crnn}. Then, for each time step $t=0,1,\ldots,T+1$, we feed $l_{1},\ldots,l_{w}\in L$ into bidirectional LSTM, with D=256 output channels per direction, which can be formulated as follows:

 \begin{equation}\label{eq10}
\begin{aligned}
h^{'}_t=f(x_t,h^{'}_{t-1})\\
y_t=\varphi (h^{'}_t)=softmax(W_0h^{'}_t)\\
\end{aligned}
\end{equation}

where $f()$ is the recurrence formulation, $h_t$ is the hidden state at time step t, and the $W_0$ linearly transforms hidden states to the output space of size 12, including 10 Arabic numerals and a token representing ``.", and a special END token. Finally, a CTC layer is applied to align the predicted sequence to the label sequence. Following \cite{crnn}, the recognition loss can be formulated as  

 \begin{equation}\label{eq11}
\begin{aligned}
L_{num\_reco}=-\frac{1}{N}\sum_{n=1}^{N}logp(y^{*}_n\mid x)
\end{aligned}
\end{equation}

where $N$ is the number of number regions in an input image, and $y^*_n$ is the recognition label.

\textbf{Training procedure and inference.}
VAM is a unified module that can be trained end-to-end. The overall loss function can be calculated as follows:

 \begin{equation}\label{eq12}
\begin{aligned}
L=L_{com}+L_{num\_det}+L_{num\_reco}
\end{aligned}
\end{equation}

In our inference process, binarized score maps for pointer and key scale are first obtained by applying the threshold algorithm $\lambda=0.5$. Then, a thinning algorithm is applied to turn the pointer into a straight line segmentation, and the Hough line transform is used to obtain the position of the pointer. Meanwhile, the key scale centers can be localized by calculating the average pixel position within the closed area. Finally, the meter reading is calculated by the angle method, which is given by

 \begin{equation}\label{eq13}
\begin{aligned}
Result=\frac{\alpha_1 }{\alpha_2} \times  num\_rec
\end{aligned}
\end{equation}

where $\alpha_1$ is the angle between the pointer and the zero scale and $\alpha_2$ is the angle between the zero scale and the key scale. The $num\_rec$ is the output of the meter number recognition branch, and then the reading of the meter is completed automatically.

\section{Experiments}
\label{sec:experiment}

\begin{figure}[t]
\begin{center}
\includegraphics[width=\linewidth]{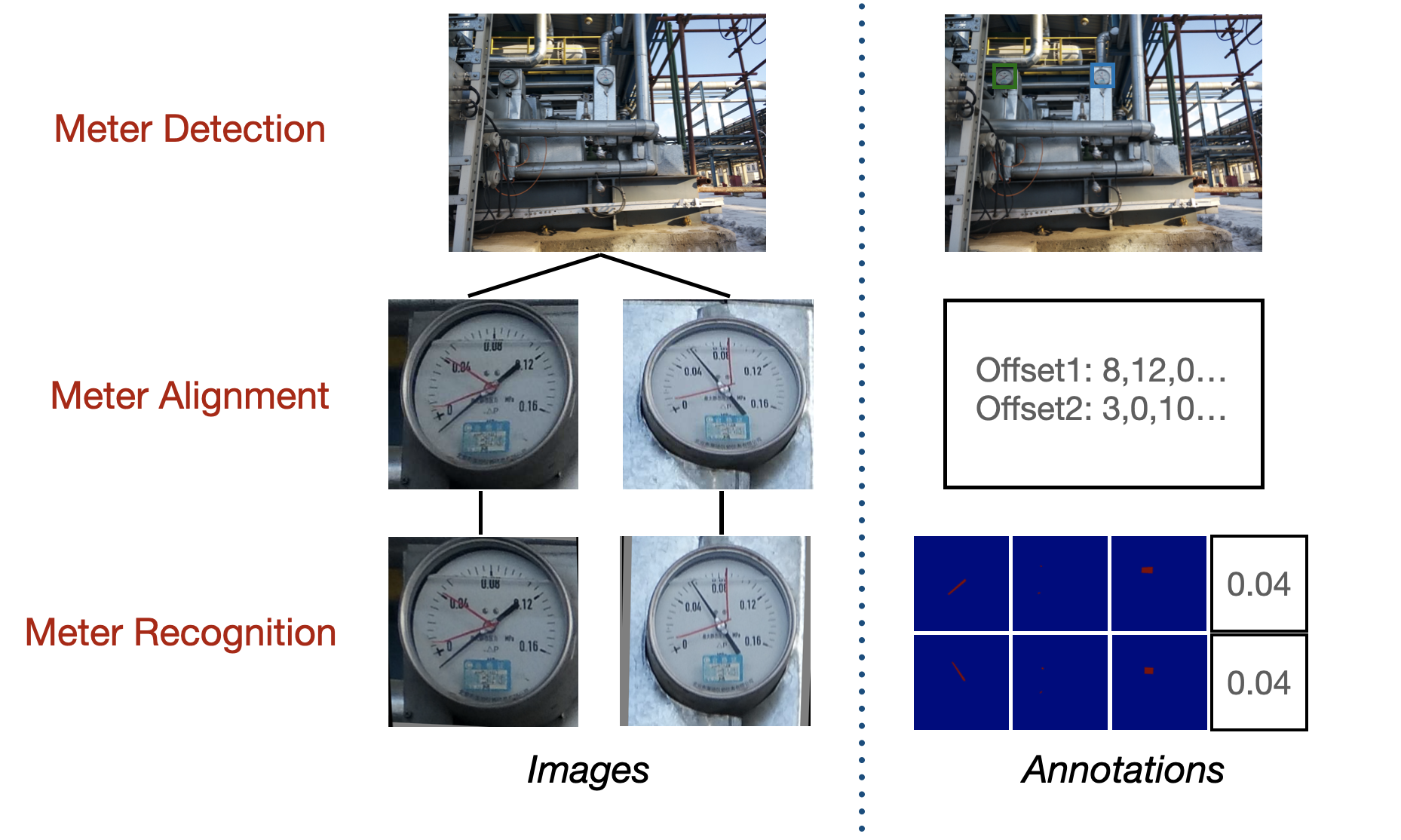}
\end{center}
\vspace{0pt}
\caption{
Visualization results of one sample in the dataset.
}
\label{figdata}
\end{figure}
\begin{figure*}[t]
\begin{center}
\includegraphics[width=\linewidth]{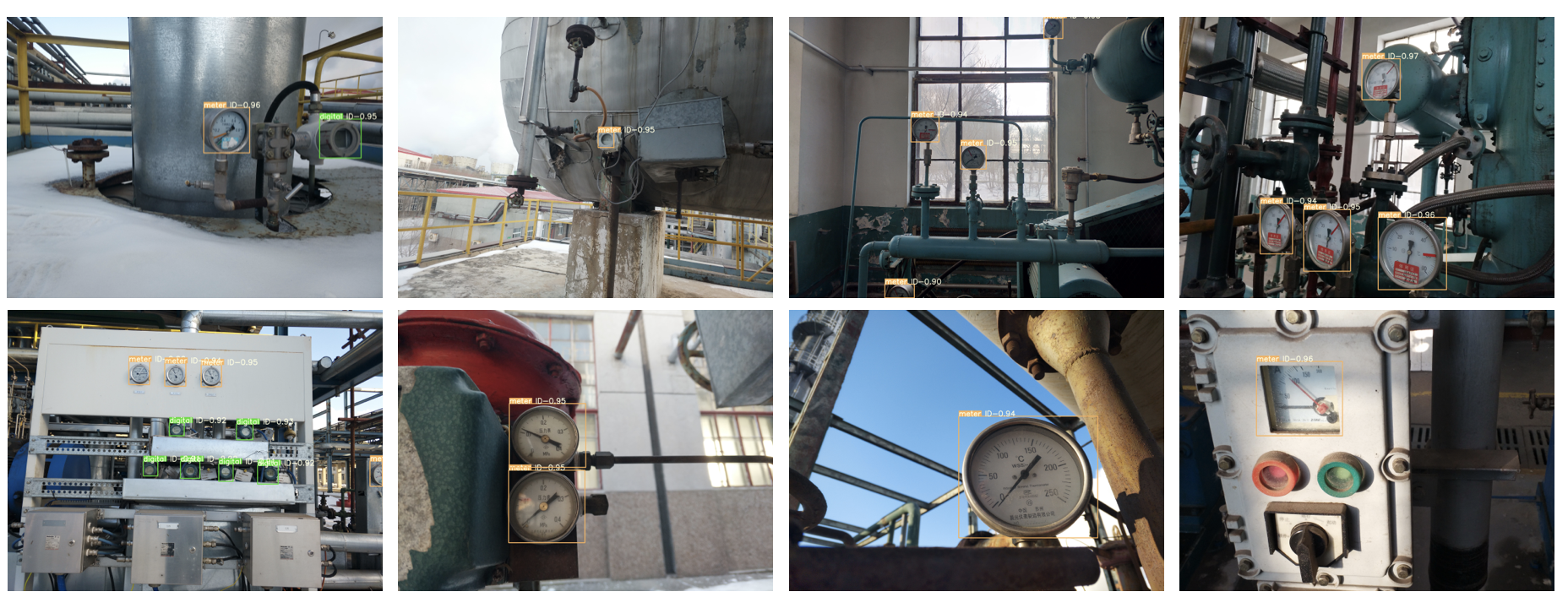}
\end{center}
\vspace{0pt}
\caption{
Qualitative results of the meter detection, where the yellow bounding box indicates the pointer meter and the green bounding box indicates the digital meter. ``ID-num" is the detection confidence.
}
\label{figdetect}
\end{figure*}

\subsection{Datasets}
To our knowledge, there have been no publicly available and appropriate benchmarks for this task. As a result, we created a new dataset called Meter Challenge (MC1296), which contains 1296 images of scenes captured by automated robots. To help the model adapt to its natural environment, the dataset includes complex backgrounds, multiple scales, a variety of viewpoint angles, and a variety of meter shapes. To better fit the meter reading task, we organized the dataset into a tree structure, with each level representing a distinct task (meter detection, meter alignment, and meter recognition), complete with associated images, annotations, and evaluation metrics. Fig \ref{figdata} illustrates some visualization results, while Table \ref{dataset} contains summary statistics.  

\begin{table}
\normalsize
\centering
\captionsetup{font=normal}
\caption{Statistics of the proposed MC 1260 dataset. ``mb",``co", and ``psn" represent the meter bounding box, coordinate offsets, and pointer/scale/number mask and number, respectively.}
{ 
\begin{tabular}{@{}lccc@{}}
\toprule
Dataset\_task & Train\_size & Test\_size &  Annotations     \\ 
\midrule
M\_detection	& 1036	& 260	& mb	 \\ 
M\_alignment	& 1028	& 247	& co	 \\ 
M\_reading	& 739	& 185	& psn	 \\ 

\bottomrule
\end{tabular}
} 

\label{dataset}
\end{table}
\begin{table}
\normalsize
\centering
\captionsetup{font=normal}
\caption{The quantitative results of different methods for meter detection.}
{ 
\begin{tabular}{@{}lccc@{}}
\toprule
Model & AP50(\%) & AP75(\%) &  FPS     \\ 
\midrule
Liu.et al \cite{meter4}	& 91.3	& 89.5	& 4.3	 \\ 
YOLO \cite{yolo}	& 90.0	& 88.2	& 6.7	 \\ 
Ours	& 98.6	& 97.1	& 12.4	 \\
\bottomrule
\end{tabular}
} 
\label{detect}
\end{table}

\subsection{Implementation details}
In this paper, the system we propose consists of YDM, STM, and VAM. YDM has similar configurations to \cite{yolov4}, so we focus on the implementation of STM and VAM. Specifically, for both of modules we use ResNet pretrained in ImageNet\cite{imagenet} as the backbone, and the image size is 640 \time 640 and the training batch size is 8. We use Adam to optimize the two networks and set the initial learning rate to $1\times 10^{-4}$ with a momentum of 0.9.   

Meanwhile, some basic data augmentation techniques are applied, such as random cropping, random rotation, and brightness contrast. Our experiment is conducted on one general GPU(GTX-1080), with the environment PyTorch 1.5.0. 

\begin{figure*}[t]
\begin{center}
\includegraphics[width=\linewidth]{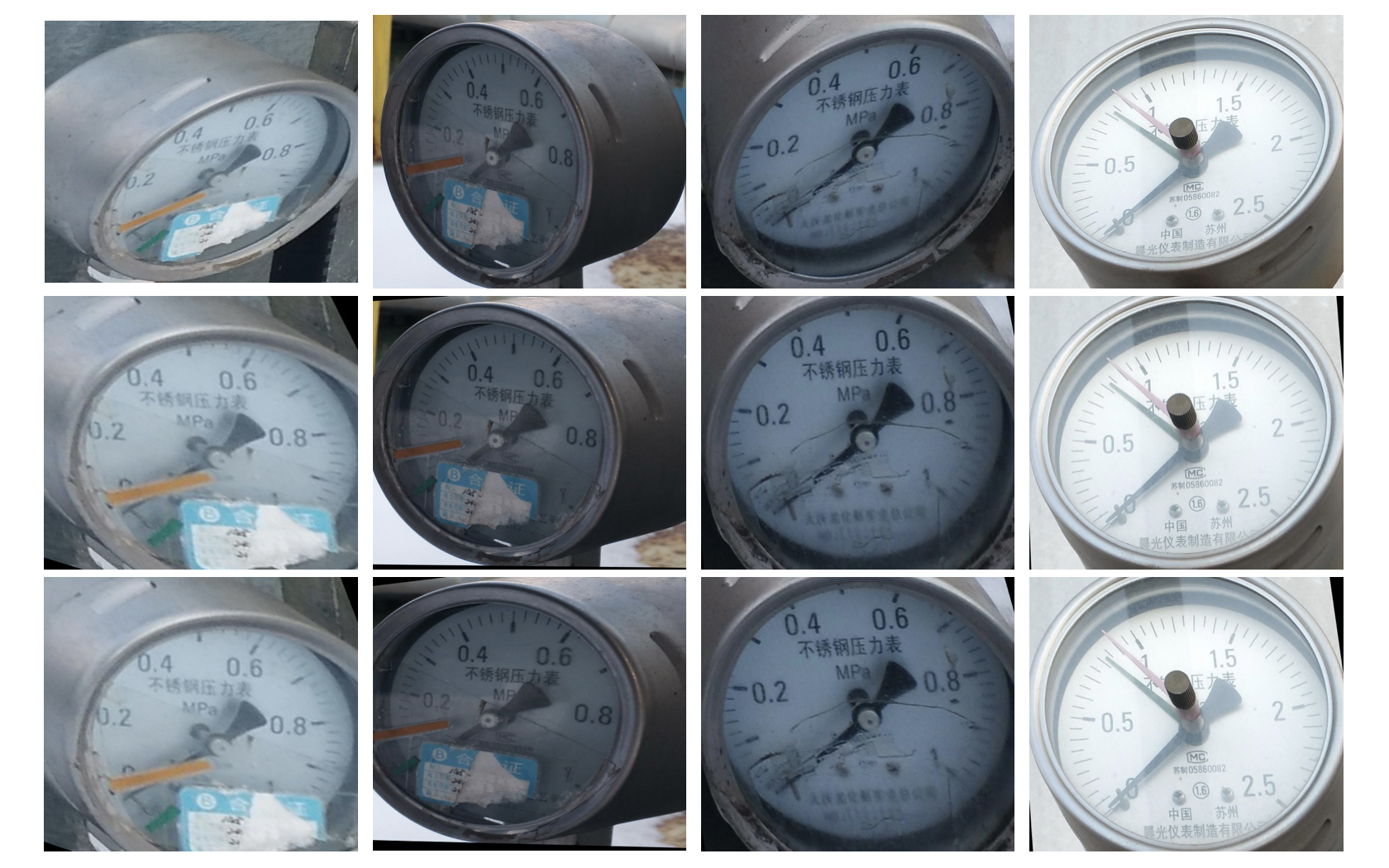}
\end{center}
\vspace{0pt}
\caption{
Qualitative results of the meter alignment, where the top row is the original images, and the middle row and the bottom row are the transformed images generated by the STN and STM. Note that the STN cannot handle images with extremely large camera angles.  
}
\label{figalign}
\end{figure*}
\subsection{Meter detection results}
To disentangle the effects of YDM, we begin by reporting the dataset’s meter detection results. To conform to the object detection literature, we report the average precision (AP) at two different bounding box IoU thresholds, AP50 and AP75. AP50 denotes the average precision for IoU thresholds greater than 0.5, while AP75 denotes the average precision for IoU thresholds greater than 0.75. As shown in Table \ref{detect}, the meter detection task is relatively successful. To demonstrate the advantages of our method, we compare it to a commonly used YOLO algorithm \cite{yolo}  and the method in \cite{meter4} , which demonstrates that our YDM performs better in terms of accuracy and efficiency. The qualitative results are demonstrated in Fig. \ref{figdetect}, which shows that TDM can detect meters with different shapes and sizes.

\begin{table}
\normalsize
\centering
\captionsetup{font=normal}
\caption{The quantitative results of different methods for meter alignment.``rel" is the average relative error, and ``ref" is the average reference error.  }
{ 
\begin{tabular}{@{}lccc@{}}
\toprule
Method & Rel(\%) & Ref(\%) & FPS     \\ 
\midrule
None	& 5.91	& 1.20	& -	 \\ 
Perspective transform \cite{meter4}	& 1.72	& 0.23	& 10	 \\ 
STN \cite{stn}	& 3.40	&  0.95	& 44	 \\
STM	& 1.70	& 0.26	& 50	 \\
\bottomrule
\end{tabular}
} 
\label{align}
\end{table}
\subsection{Meter alignment results}
To demonstrate the STM’s availability and robustness in the recognition system, we conducted extensive experiments on the validation dataset, comparing it to the traditional perspective transform method and STN. Fig. \ref{figalign} illustrates the qualitative findings. As seen, the image can be easily and automatically transformed into a front-viewing image using STM, regardless of the camera angle. However, due to the limited learning capability of pure STN, it is difficult to align the meter in terms of some extremely large camera angles.

Additionally, as shown in Table \ref{align}, we conducted ablation studies to demonstrate its superiority by demonstrating inference speed and influence on the meter recognition task. Note that the average relative error and the average reference error are the evaluation metrics used to represent the meter recognition error rate, which will be discussed in detail in Sec \ref{recognition_result}. It can be seen that STM contributes to reducing the recognition error rate, as it allows meters to be read from various angles and sizes. Our STM also achieves competitive accuracy to perspective transform while increasing inference speed, indicating that STM achieves a more favorable trade-off between accuracy and efficiency.

\begin{table}
\normalsize
\centering
\captionsetup{font=normal}
\caption{The quantitative results of different methods for meter reading recognition.``Rel" is the average relative error, and ``Ref" is the average reference error.  }
{ 
\begin{tabular}{@{}lccc@{}}
\toprule
Method & Avenue  & Rel(\%) & Ref(\%)      \\ 
\midrule
Zheng et al. \cite{meter2} & Measurement(2016) & 10.32	& 0.91	 \\ 
Gao et al. \cite{article2} & ICRAS(2017) & 9.34	& 0.67	 \\ 
He et al. \cite{metermaskrcnn}	& ICIST(2019)	& 1.85	& 0.30	 \\
Liu et al. \cite{meter4}	& Measurement(2020)	&  1.77	& \textbf{0.24}	 \\
Ours	& -	& \textbf{1.70}	& 0.26	 \\
\bottomrule
\end{tabular}
} 
\label{result}
\end{table}
\begin{figure*}[t]
\begin{center}
\includegraphics[width=\linewidth]{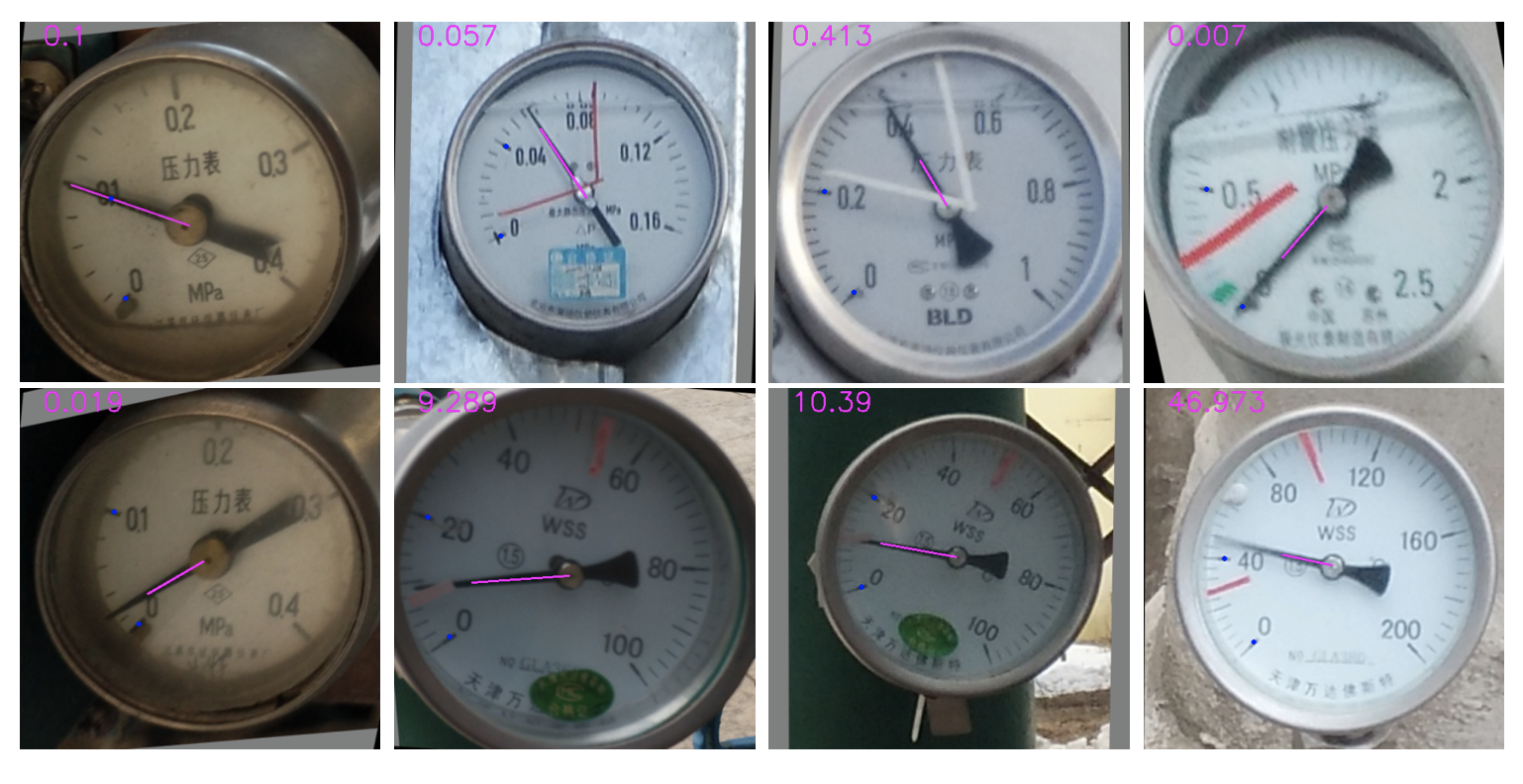}
\end{center}
\vspace{0pt}
\caption{
Some visualization results produced by our method. The red line is the predicted pointer line, the blue points are the key scale areas, and the meter reading results are shown in the top left.
}
\label{figresult}
\end{figure*}
\subsection{Meter recognition results}
\label{recognition_result}
To demonstrate our method's recognition performance, we incrementally compare it to other methods. To minimize interperson variability in readings, the readings obtained by human vision are the average of the results of twenty expert workers. Meanwhile, to make the comparison fairer, we follow similar evaluation metrics as \cite{metermaskrcnn}. Specifically, we choose the average relative error $\hat{\Theta}$ and the average reference error $\hat{\Gamma}$ as evaluation indicators, as shown in the following:

 \begin{equation}\label{eq14}
\begin{aligned}
\hat{\Theta}=\frac{\sum_{i=1}^{n}\frac{\mid p_i-g_i \mid}{g_i}}{n} \times 100\% \\
\hat{\Gamma}=\frac{\sum_{i=1}^{n}\frac{\mid p_i-g_i \mid}{R}}{n} \times 100\% 
\end{aligned}
\end{equation}

where $p_i$ is the predicted meter value, and $g_i$ is the ground truth value. R represents the meter`s range, and n represents the total number of experimental data. As shown in Table \ref{result}, our method outperforms previous methods in terms of average relative error and achieves competitive results with \cite{meter4} in average reference error, indicating that our algorithm has a strong capacity in reading recognition. Additionally, our method can perform inference at a rate of approximately 25 frames per second, demonstrating that it is practical for real-world applications. We show some visualization results in Fig \ref{figresult}, demonstrating our method's high adaptability to a complex environment with variable illumination, scale, and image tilt.

To disentangle the effects of the unified framework VAM, we conduct ablation studies to investigate the relationships between the meter component retrieval and meter numbering recognition branches. We begin by reporting the full model's end-to-end results in Table \ref{ram}. Notably, we evaluate pointer/key scale detection and key number recognition using the AP50 and number-level accuracy recognition metrics, respectively. It can be demonstrated that by optimizing all loss functions simultaneously, our model achieves a reasonable level of success in detection and recognition tasks. Additionally, we construct a two-stage model in which the meter component retrieval and meter number recognition branches are trained independently. The meter component retrieval network is built by removing the meter number recognition branch, and similarly, the meter number recognition network is built by removing the meter component retrieval branch from the original network. Our proposed VAM outperforms the two-stage method by a significant margin in both the meter component retrieval and meter number recognition tasks. The results indicate that our joint training strategy accelerated the convergence of the model parameters.     
\begin{table}
\normalsize
\centering
\captionsetup{font=normal}
\caption{Ablation studies on VAM. ``MCRB" and ``MNRB" represent that we only train the meter component retrieval branch or meter number recognition branch.}
\begin{tabular}{@{}lccc@{}}
\toprule
Method & pointer\_det(\%)  & key scale\_det(\%)  & key number\_reco(\%)  \\ 
\midrule
VAM & 95.6 & 93.2	& 88.7	 \\ 
MCRB & 93.1 & 90.5	& -	 \\ 
MNRB	& -	& -	& 87.2	 \\
\bottomrule
\end{tabular}
\label{ram}
\end{table}

\section{Conclusion}
\label{sec:conclusion}
We propose a novel method for accurate and efficient pointer meter reading, which is implemented by the YDM, STM, and VAM equipment. Specifically, STM can obtain the front view of images autonomously with the improved STN, and VAM can recognize meters accurately with unified frameworks with the combination of the meter component retrieval branch and meter number recognition branch. Experiments on the challenging datasets we proposed demonstrate that the proposed method has a strong capacity for pointer meter reading. Currently, the algorithm has been successfully applied to robots performing substation inspections. Future work will concentrate on model acceleration to develop a more efficient framework for video meter reading.

\section{Funding}
\label{sec:ack}
This work was supported in part by the National Key Research and Development Program of China under Grants 2020YFB1406902 and A12003.

%
%
%
\bibliographystyle{splncs04}
\bibliography{reference}

\end{document}